# ROBUST ADAPTIVE FUZZY SLIDING MODE CONTROL
# FOR RAJECTORY TRACKING FOR OF CYLINDRICAL MANIPULATOR

**Van Cuong Pham[1,2], Minh Hai Tran[2], Phuc Anh Nguyen[1],**
**Ngoc Son Vu[1*] , Nga Nguyen Thi[2]**
[1] Hanoi University of Industry, Hanoi, Vietnam

[2] Thai Binh University, Thai Binh Province, Vietnam
* Email: vungocson.haui.271199@gmail.com

**Abstract:** This research proposes a robust adaptive fuzzy sliding mode control (AFSMC) approach to enhance the trajectory tracking performance of cylindrical robotic manipulators, extensively utilized in applications such as CNC and 3D printing. The proposed approach integrates fuzzy logic with sliding mode control (SMC) to bolster adaptability and robustness, with fuzzy logic approximating the uncertain dynamics of the system, while SMC ensures strong performance. Simulation results in MATLAB/Simulink demonstrate that AFSMC significantly improves trajectory tracking accuracy, stability, and disturbance rejection compared to traditional methods. This research underscores the effectiveness of AFSMC in controlling robotic manipulators, contributing to enhanced precision in industrial robotic applications.

**Keywords:** Adaptive Fuzzy Sliding Mode Control (AFSMC), Sliding Mode Control (SMC), Fuzzy Logic, Robotic Manipulators, Cylindrical Manipulator

## 1. INTRODUCTION

Cylindrical robotic manipulators, combining a prismatic and a revolute joint, are extensively utilized in applications such as CNC machining, 3D printing, and assembly tasks. Precise trajectory tracking remains challenging due to nonlinear dynamics, uncertainties, and external disturbances [1].

Sliding mode control (SMC) is a well-established robust control strategy extensively applied in various fields, including robotic manipulators, electric motors, and power converters, to achieve precise trajectory tracking under uncertain and dynamic conditions [2-5]. Its ability to provide high-performance control makes it an attractive option; however, traditional SMC methods often face critical challenges such as chattering effects and sensitivity to parameter variations, which can negatively impact system stability and practical implementation [6-10]. The presence of chattering not only reduces control accuracy but may also cause mechanical wear and energy inefficiency, making it necessary to explore improved strategies for enhancing SMC performance.

To address the limitations of traditional SMC, researchers have integrated fuzzy logic control (FLC) to develop adaptive fuzzy sliding mode control (AFSMC) [11-13]. This method utilizes FLC's inference capabilities to dynamically adjust SMC parameters, enhancing robustness against uncertainties and nonlinear disturbances while reducing chattering. AFSMC allows real-time tuning, making it suitable for systems with varying operating conditions. In this research, a robust AFSMC scheme is proposed for cylindrical manipulators to improve trajectory tracking accuracy in complex environments. MATLAB/Simulink simulations demonstrate the method's effectiveness, showing significant improvements in both stability and precision compared to conventional approaches.

The research is structured as follows: Section 2 presents the manipulator's dynamic model; Section 3 details the sliding mode controller design; Section 4 provides simulation results and Section 5 concludes the study with future work.

## 2. DYNAMIC OF CYLINDRICAL MANIPULATOR

The dynamics of cylindrical manipulator under external disturbances can be described using the Lagrangian formulation as follows [1]:

$$M(q)\ddot{q} + C(q,\dot{q})\dot{q} + F(q)q + G(q) = \tau - f_{ext} \qquad (1)$$

where: $(q, \dot{q}, \ddot{q}) \in R^{n \times 1}$ is the vectors of joint positions, velocities, and accelerations. $M(q) \in R^{n \times n}$ is the symmetric inertial matrix. $C(q) \in R^{n \times n}$ is the vector of coriolis and centripetal forces. $G(q) \in R^{n \times 1}$ is the gravity vector. $F(q)$ represents the vector of the frictions. $f_{ext} \in R^{n \times 1}$ is the unknown disturbances input vector. $\tau \in R^{n \times 1}$ is the joints torque input vector.

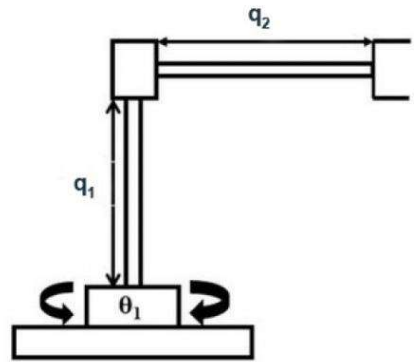

**Figure 1.** The cylindrical manipulator

In [1], writing equations dynamics of a cylindrical manipulator in the standard matrix form, we have:

$$\begin{bmatrix} A_{11} & A_{12} & A_{13} \\ A_{21} & A_{22} & A_{23} \\ A_{31} & A_{32} & A_{33} \end{bmatrix} \times \begin{bmatrix} \ddot{\theta}_1 \\ \ddot{q}_2 \\ \ddot{q}_3 \end{bmatrix} + \begin{bmatrix} B_{11} & B_{12} & B_{13} \\ B_{21} & B_{22} & B_{23} \\ B_{31} & B_{32} & B_{33} \end{bmatrix} \times \begin{bmatrix} \dot{\theta}_1{}^2 \\ \dot{q}_2{}^2 \\ \dot{q}_3{}^2 \end{bmatrix} + \begin{bmatrix} C_{11} & C_{12} & C_{13} \\ C_{21} & C_{22} & C_{23} \\ C_{31} & C_{32} & C_{33} \end{bmatrix} \times$$

$$\begin{bmatrix} \dot{\theta}_1 \dot{q}_2 \\ \dot{\theta}_1 \dot{q}_3 \\ \dot{q}_2 \dot{q}_3 \end{bmatrix} + \begin{bmatrix} D_1 \\ D_2 \\ D_3 \end{bmatrix} = \begin{bmatrix} \tau_1 \\ \tau_2 \\ \tau_3 \end{bmatrix} \tag{2}$$

In which:

$A_{11} = (4m_1 \sin\theta_1 - 4m_2 \cos\theta_1)q_3 + I_3$; $A_{13} = (m_1 + m_2)(\sin\theta_1 \cos\theta_1)q_3$;

$A_{22} = m_3$; $A_{31} = m_1 \sin\theta_1 \cos\theta_1$; $A_{33} = 2(m_1 \sin\theta_1 + m_2 \cos\theta_1)$;

$A_{12} = A_{21} = A_{23} = A_{32} = 0$;

$B_{11} = (m_1 \sin\theta_1 - 4m_2 \cos\theta_1)q_3$; $B_{13} = -m_1 \cos\theta_1 + m_2 \sin\theta$;

$B_{31} = 2q_3(m_1 \sin\theta_1 - m_2 \cos\theta_1)$; $B_{12} = B_{21} = B_{22} = B_{23} = B_{32} = 0$

$C_{12} = -(m_1 + m_2)(\sin\theta_1 \cos\theta_1)q_3$; $C_{32} = -(m_1 + m_2)(\sin\theta_1 \cos\theta_1)$;

$C_{11} = C_{13} = B_{21} = C_{22} = C_{23} = C_{31} = 0$;

$D_2 = g(m_2 + m_3)$; $D_1 = D_3 = 0$

in which $m_i$ are the mass of joint of Cylindrical Manipulator, respectively, $I_3 = 1\left(\frac{kg}{m^2}\right)$ is the moment of inertia of joint 3, respectively, and $g = 9.8\left(\frac{m}{s^2}\right)$ is acceleration of gravity. The values of the parameters of each joint of the manipulator are shown in Table 1.

*Table 1.* Values of masses (m) of each joint [1]

| Joint | m (kg) |
|---|---|
| 1 ($\theta_1$) | 36.367405 |
| 2 ($q_2$) | 12.632222 |
| 3 ($q_3$) | 23.735183 |

The position trajectories of the Cylindrical Manipulator are chosen: $[\theta_1 \quad q_2 \quad q_3]_0^T = [0.01 \quad 0.01 \quad 0.01]^T$ and initial velocities of joints are $[\dot{\theta}_1 \quad \dot{q}_2 \quad \dot{q}_3]_0^T = [0 \quad 0 \quad 0]^T$

## 3. AFSMC CONTROLLER DESIGN

Consider the trajectory tracking errors defined as: $e(t) = q_d(t) - q(t)$; $\dot{e}(t) = \dot{q}_d(t) - \dot{q}(t)$. Where $q(t)$ and $\dot{q}(t)$ is the actual joint position and velocity.

The sliding surface is defined as: $s(t) = \dot{e}(t) + \lambda e(t)$, with $\lambda \in R^{n \times n}$ is a positive definite matrix.

The conventional sliding mode control law is expressed as:

$$\tau_{SMC} = M(q)[\ddot{q}_d - \lambda \dot{e}] + C(q, \dot{q})\dot{q} + F(q) + G(q) + K.sign(s) \tag{3}$$

Given that the cylindrical manipulator is a nonlinear system with uncertainties and disturbances, the choice of matrix $\lambda \in R^{n \times n}$ directly affects the convergence rate and robustness. Instead of using a fixed $\lambda$, a fuzzy logic is proposed to adaptively tune $\lambda(t)$ based on system errors.

- The fuzzy system uses position error $e(t)$ and velocity error $\dot{e}(t)$ as inputs, reflecting the deviation between the desired and actual trajectories.

| $\dot{e}(t)$ \ $\dot{e}(t)$ | NL | NS | ZE | PS | PL |
|---|---|---|---|---|---|
| NL | L | L | ML | M | MS |
| NS | L | ML | M | MS | S |
| ZE | ML | M | M | M | ML |
| PS | MS | MS | M | ML | L |
| PL | S | MS | ML | L | L |

**Figure 2.** Fuzzy logic law

Each input variable has five fuzzy sets: Negative Large (NL), Negative Small (NS), Zero (ZE), Positive Small (PS), Positive Large (PL)

- The output: $\lambda(t) = \lambda_1(t), \lambda_2(t), \lambda_3(t)$ also has five fuzzy sets: Small (S), Medium Small (MS), Medium (M), Medium Large (ML), and Large (L).

- The fuzzy rules are of the form: IF $e$ is [A] AND $\dot{e}$ is [B] THEN $\lambda$ is [C]

Defuzzification is performed using the WTAVER method:

$$\lambda(t) = \frac{\sum_{i=1}^{25} \bar{\lambda}_i . \mu_i(e) . \mu_i(\dot{e})}{\sum_{i=1}^{25} \mu_i(e) . \mu_i(\dot{e})} \tag{4}$$

Where: $\bar{\lambda}_i$ is the output value of the $i$, and $\mu_i(e), \mu_i(\dot{e})$ are the membership values of the inputs.

To analyze stability, consider the Lyapunov function:

$$V(t) = \frac{1}{2} s^T(t) M s(t) \tag{5}$$

Taking its time derivative:

$$\dot{V}(t) = s^T \dot{M} s + \frac{1}{2} s^T \dot{M} s \tag{6}$$

Using system dynamics and the sliding surface definition:

$$\dot{M}s = M(\ddot{e} + \dot{\lambda}e + \lambda\dot{e}) = M(\ddot{q}_d - \ddot{q} + \dot{\lambda}e + \lambda\dot{e}) \tag{7}$$

Thus: $\tau_{SMC} = \text{M}(q)[\ddot{q}_d - \lambda\dot{e}] + C(q,\dot{q})\dot{q} + F(q) + G(q) + K.sign(s)$

Substituting the control law: $\dot{M}s = -K.sign(s) + \dot{\lambda}Me.$

Thus: $\dot{V} = s^T(-K.sign(s) + \dot{\lambda}Me) + \frac{1}{2}s^T\dot{M}s.$ Given that $s^T sign(s) = \|s\|$, thus:

$$\dot{V} = -\sum_{i=1}^{n} K_i|s_i| + s^T\dot{\lambda}Me + \frac{1}{2}s^T\dot{M}s \tag{8}$$

Assuming: $\|\dot{M}\| \le \beta_1$ with $\beta_1 > 0$; $\|\dot{\lambda}\| \le \beta_2$ with $\beta_2 > 0$

$$\dot{V} = -\sum_{i=1}^{n} K_i|s_i| + \beta_2\|s\|\|M\|\|e\| + \frac{1}{2}\beta_1\|s\|^2 \tag{9}$$

To ensure asymptotic stability, select $K_i$ such that:

$$\sum_{i=1}^{n} K_i|s_i| > \beta_2\|s\|\|M\|\|e\| + \frac{1}{2}\beta_1\|s\|^2 \tag{10}$$

This ensures $\dot{V} < 0$ for all $s \ne 0$, confirming system stability via Lyapunov theory. The fuzzy logic-based adaptive tuning enhances both robustness and adaptability under dynamic uncertainties.

## 4. THE SIMULATION RESULTS

- Constant Desired Trajectory

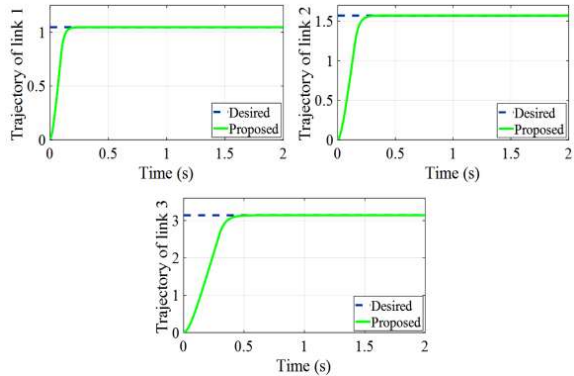

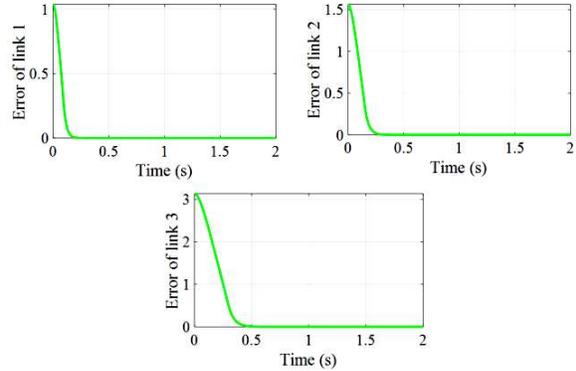

**Figure 3.** Trajectory responses of joints for the constant desired trajectory

**Figure 4.** Tracking errors of joints for the constant desired trajectory

The desired joint values were set as: $\theta_1 = \frac{\pi}{3}, q_2 = \frac{\pi}{2}, q_3 = \pi$. Simulation results show that the ASMC controller ensures fast and accurate tracking, with errors approaching zero in under 0.5 seconds. There is no overshoot, confirming good control stability. Figures 2 and 3 depict the joint responses and tracking errors, demonstrating the system's effectiveness in following the set trajectory precisely.

- Uncertain Desired Trajectory

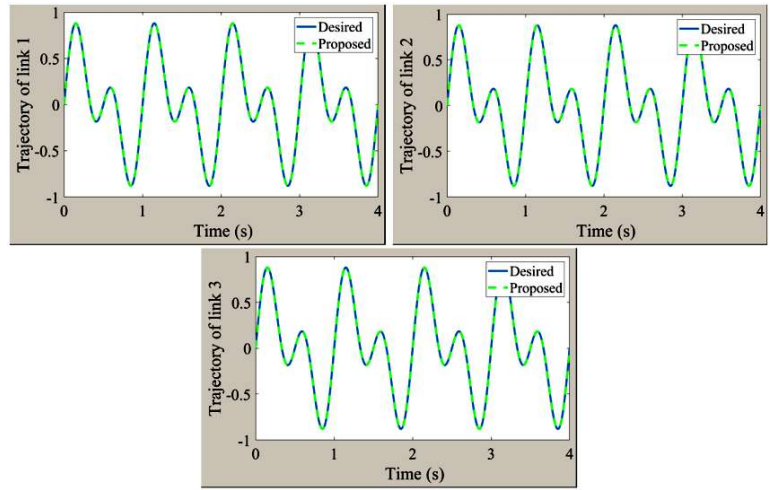

**Figure 5.** Trajectory responses of joints for the uncertain desired trajectory

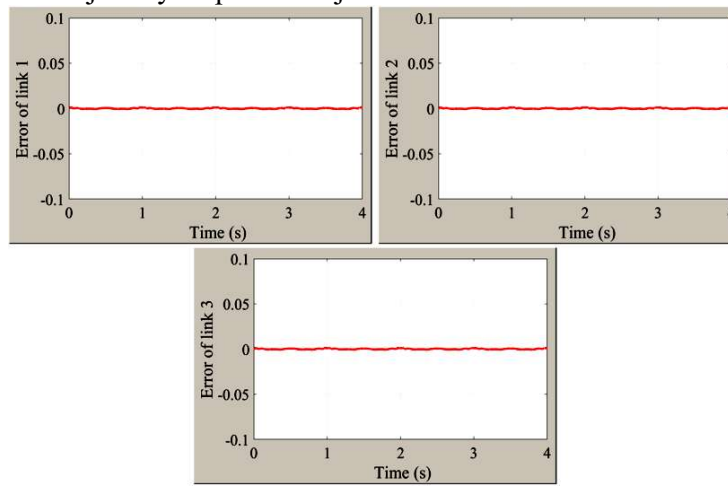

**Figure 6.** Tracking errors of joints for the uncertain desired trajectory

For an uncertain desired trajectory, the ASMC continues to perform reliably. The system maintains low error and high stability despite uncertainties and disturbances. As shown in Figures 5 and 6, the controller enables smooth and precise motion, proving the robustness and adaptability of the proposed approach.

## 5. CONCLUSION

This paper presented an Adaptive Fuzzy Sliding Mode Control (AFSMC) approach for trajectory tracking of cylindrical manipulators. By integrating fuzzy logic with sliding mode control, the proposed method improves adaptability to uncertainties and enhances robustness against external disturbances. MATLAB/Simulink simulations demonstrate that the AFSMC approach achieves better trajectory accuracy, reduced tracking errors, and improved stability compared to conventional control methods. These results highlight the effectiveness of AFSMC for precise motion control. Future research will focus on experimental validation and real-time implementation in industrial applications to further assess its performance and practicality.

## REFERENCE

[1] Batista, J., Souza, D., Dos Reis, L., Barbosa, A., & Araújo, R. (2020). Dynamic model and inverse kinematic identification of a 3-DOF manipulator using RLSPSO. Sensors, 20(2), 416.

[2] Komurcugil, H., Biricik, S., Bayhan, S., & Zhang, Z. (2020). Sliding mode control: Overview of its applications in power converters. IEEE Industrial Electronics Magazine, 15(1), 40-49.

[3] Shtessel, Y., Fridman, L., & Plestan, F. (2016). Adaptive sliding mode control and observation. International Journal of Control, 89(9), 1743-1746.

[4] Shtessel, Y., Fridman, L., & Plestan, F. (2016). Adaptive sliding mode control and observation. International Journal of Control, 89(9), 1743-1746.

[5] Utkin , Poznyak, Orlov, & Polyakov, A. (2020). Conventional and high order sliding mode control. Journal of the Franklin Institute, 357(15), 10244-10261.

[6] Nadda, S., & Swarup, A. (2018, February). Integral sliding mode control for position control of robotic manipulator. In 2018 5th International Conference on Signal Processing and Integrated Networks (SPIN) (pp. 639-642). IEEE.

[7] Islam, S., & Liu, X. P. (2010). Robust sliding mode control for robot manipulators. IEEE Transactions on industrial electronics, 58(6), 2444-2453.

[8] Adhikary, N., & Mahanta, C. (2018). Sliding mode control of position commanded robot manipulators. Control Engineering Practice, 81, 183-198.

[9] Baek, J., Kwon, W., & Kang, C. (2020). A new widely and stably adaptive sliding-mode control with nonsingular terminal sliding variable for robot manipulators. IEEE Access, 8, 43443-43454.

[10] Wang, Y., Zhang, Z., Li, C., & Buss, M. (2022). Adaptive incremental sliding mode control for a robot manipulator. Mechatronics, 82, 102717.

[11] Fateh, S., & Fateh, M. M. (2023). Superior Adaptive Fuzzy Sliding Mode Control of Electrically Driven Robot Manipulators. Iranian Journal of Science and Technology, Transactions of Electrical Engineering, 47(2), 491-502.

[12] Li, Z., & Zhai, J. (2024). Fuzzy adaptive super-twisting sliding mode asymptotic tracking control of robotic manipulators. International Journal of Fuzzy Systems, 26(1), 34-43.

[13] Thinh Ngo, H. Q., Shin, J. H., & Kim, W. H. (2008). Fuzzy sliding mode control for a robot manipulator. Artificial Life and Robotics, 13, 124-128.

# ĐIỀU KHIỂN TRƯỢT MỜ BỀN VỮNG THÍCH NGHI CHO TAY MÁY ROBOT CYLINDRICAL

**Van Cuong Pham[1,2], Minh Hai Tran[2], Phuc Anh Nguyen[1],
Ngoc Son Vu[1*] , Nga Nguyen Thi[2]**

[1] *Hanoi University of Industry, Hanoi, Vietnam*

[2] *Thai Binh University, Thai Binh Province, Vietnam*

*\* Email: vungocson.haui.271199@gmail.com*

**Tóm tắt:** Nghiên cứu này đề xuất một phương pháp điều khiển trượt mờ thích nghi bền vững (AFSMC) nhằm cải thiện hiệu suất bám sát quỹ đạo của tay máy robot, được sử dụng rộng rãi trong các ứng dụng như máy CNC và in 3D. Phương pháp đề xuất kết hợp logic mờ với điều khiển trượt để tăng cường tính thích nghi và độ bền vững, trong đó logic mờ giúp xấp xỉ các động học bất định của hệ thống, còn SMC đảm bảo hiệu suất mạnh mẽ. Kết quả mô phỏng trên MATLAB/Simulink cho thấy AFSMC cải thiện đáng kể độ chính xác bám sát quỹ đạo, nâng cao sự ổn định và khả năng chống nhiễu so với các phương pháp truyền thống. Nghiên cứu này nhấn mạnh tính hiệu quả của AFSMC trong việc điều khiển tay máy robot, góp phần nâng cao độ chính xác trong các ứng dụng robot công nghiệp.

**Từ khóa:** Điều khiển trượt mờ thích nghi (AFSMC), Điều khiển trượt (SMC), Logic mờ, tay máy robot